\documentclass[11pt]{article}

\usepackage[letterpaper,margin=1in]{geometry}
\usepackage[utf8]{inputenc}
\usepackage[T1]{fontenc}
\usepackage{lmodern}
\usepackage[hyphens]{url}
\usepackage{graphicx}
\usepackage[numbers,sort&compress]{natbib}
\usepackage{caption}
\usepackage{microtype}
\usepackage[table]{xcolor}
\usepackage{soul}
\usepackage{hyperref}
\hypersetup{
  colorlinks=true,
  linkcolor=blue!55!black,
  citecolor=green!40!black,
  urlcolor=blue!60!black,
  pdfauthor={Xuyang Cao, Enyou Liu, Jun Zhao, Zhuoyun Liu, Jintao Fei, Leo},
  pdftitle={SAM Meets VLM: Parameter-Decoupled Full-Parameter Training for Unified Medical Reasoning and Segmentation}
}
\usepackage{amsmath}
\usepackage{amssymb}
\usepackage{array}
\usepackage{booktabs}
\usepackage{multirow}
\usepackage{algorithm}
\usepackage{algorithmic}

\newif\ifshowrevisions
\showrevisionsfalse
\sethlcolor{yellow!45}
\newcommand{\revtext}[1]{\ifshowrevisions\hl{#1}\else#1\fi}
\newcommand{\revmath}[1]{%
  \ifshowrevisions\colorbox{yellow!45}{\ensuremath{\displaystyle #1}}%
  \else\ensuremath{#1}\fi}
\newcommand{\revcell}[1]{\ifshowrevisions\cellcolor{yellow!45}#1\else#1\fi}

\title{SAM Meets VLM: Parameter-Decoupled Full-Parameter Training for Unified Medical Reasoning and Segmentation}

\author{%
  Xuyang Cao\textsuperscript{*} \quad
  Enyou Liu\textsuperscript{*} \quad
  Jun Zhao \quad
  Zhuoyun Liu \quad
  Jintao Fei \quad
  Leo\\[0.5em]
  \normalsize JDH Algo, JD Health International Inc.\\
  \small \textsuperscript{*}Equal contribution.
}
\date{}

\begin{document}
\maketitle

\begin{abstract}
Medical multimodal large language models (MLLMs) are increasingly expected
not only to answer clinical questions, but also to localize the visual
evidence behind their predictions.
A common strategy connects a vision--language model (VLM) with SAM-style
segmentation through a special \texttt{<SEG>} token, yet full-parameter
training of this unified architecture is difficult because image-level
reasoning and pixel-level segmentation impose different requirements on the
shared representation space.
To address this issue, we propose a parameter-decoupled training framework
for unified medical reasoning and segmentation.
The framework treats the \texttt{<SEG>} hidden state as a semantic-to-spatial
prompt for the mask decoder and encourages it to become separable from
generic language states, reducing ambiguous segmentation prompts and
potential disruption to reasoning representations.
It first performs medical shallow alignment to adapt visual features to
clinical language without disturbing the LLM; then controlled instruction
tuning shapes separable \texttt{<SEG>} prompt states, monitored by the
Davies--Bouldin Index (DBI), while scaling segmentation gradients entering
the language backbone; finally, the SAM branch is specialized with the VLM
frozen to improve mask precision without altering reasoning parameters.
Experiments on medical referring segmentation, grounding, visual QA, and textual QA
benchmarks show that our framework achieves strong language-conditioned
segmentation while preserving competitive reasoning ability.
Ablations show that two-phase instruction tuning, gradient scaling, and
segmentation specialization all contribute to the model.
\end{abstract}

\section{Introduction}

Recent studies show that augmenting vision--language models (VLMs)
with segmentation modules enables medical multimodal large language models
(MLLMs) to perform not only image-level understanding, such as diagnosis and
report generation, but also pixel-level reasoning, including the localization
of lesions or anatomical structures~\cite{wu2025unibiomed,chen2025mimo}.
This capability matters for clinical use cases in which an answer is
more useful when accompanied by spatial evidence.
However, training a single model that supports both medical reasoning and
language-conditioned segmentation remains challenging.
Image-level reasoning benefits from high-level semantic abstraction, whereas
segmentation requires fine-grained spatial cues and stable prompts for a mask
decoder.
When these objectives are optimized together in the same model, the shared
representations can be pulled toward incompatible levels of visual detail.

Inspired by LISA~\cite{lai2024lisa} and GLaMM~\cite{rasheed2024glamm},
recent medical MLLMs couple SAM~\cite{kirillov2023segment,ravi2024sam} with
VLMs~\cite{liu2023visual,yang2025qwen2.5} through a special
\texttt{<SEG>} token whose hidden state is projected as an implicit
segmentation prompt.
MIMO~\cite{chen2025mimo} and UniBioMed~\cite{wu2025unibiomed} follow this
paradigm with adapter-based tuning~\cite{hu2022lora}, while MedPLIB~\cite{huang2024towards}
uses separate prediction heads.
These designs demonstrate the promise of unified medical interpretation, but
they leave open a key training question: how can a full-parameter medical
MLLM learn expressive segmentation prompts without eroding its medical
vision--language reasoning ability?

We address this question with a parameter-decoupled training framework.
Instead of treating the \texttt{<SEG>} token as a passive interface, we
explicitly monitor and regularize its hidden-state distribution so that it
becomes a stable, SAM-addressable prompt space.
The curriculum separates shallow medical alignment, controlled
instruction tuning, and segmentation specialization.
This design is motivated by empirical training interference observed during
naive joint optimization, but the paper's main contribution is the practical
training recipe and its validation across reasoning and segmentation tasks.

\noindent\textbf{Our contributions are summarized as follows:}
\begin{itemize}
  \item We study full-parameter unified medical MLLM training for joint
        medical reasoning and language-conditioned segmentation, and analyze
        the representation-level interference that arises around the
        \texttt{<SEG>} interface.
  \item We propose a three-stage parameter-decoupled training framework with
        DBI-monitored \texttt{<SEG>} state separation, module-specific
        gradient scaling, and VLM-frozen SAM specialization.
  \item We evaluate the model on segmentation, grounding, medical VQA, and
        textual QA benchmarks, and provide ablations showing how each
        training component affects the reasoning--segmentation trade-off.
\end{itemize}

\section{Related Work}

\subsection{Medical Multimodal Large Language Models}
Building on general-domain VLMs~\cite{liu2023visual,yang2025qwen2.5}
and vision--language pre-training such as
CLIP~\cite{radford2021clip} and MedKCO~\cite{zhang2026medkco},
recent medical MLLMs integrate clinical knowledge for tasks such as
report generation, visual question answering, and image
interpretation.
Early efforts including LLaVA-Med~\cite{li2023llavamed},
RadFM~\cite{wu2023radfm}, and
PMC-LLaMA~\cite{wu2024pmcllama} established strong medical
vision--language foundations through large-scale biomedical
pretraining.
Lingshu~\cite{xu2025lingshu} and HealthGPT~\cite{lin2025healthgpt}
show strong performance on medical VQA through large-scale medical
pretraining.
MedGemma~\cite{sellergren2025medgemma} leverages Google's Gemma
backbone for diverse medical modalities.
HuatuoGPT-V~\cite{chen2024towards} focuses on injecting medical
visual knowledge at scale.
However, these models focus on image-level understanding and do not
support pixel-level segmentation.
MIMO~\cite{chen2025mimo} and UniBioMed~\cite{wu2025unibiomed} extend
this paradigm by coupling SAM-based segmentation with VLMs, but rely
on LoRA-based tuning~\cite{hu2022lora} to avoid gradient interference;
Sparse Spectral LoRA~\cite{nejatimanzari2026sparse} further studies
routed adapter experts for robust medical VLM adaptation.
MedPLIB~\cite{huang2024towards} employs a mixture-of-experts
design and explicit prediction-head decoupling.
Our work instead focuses on the full-parameter training regime and studies
how to preserve a segmentation-oriented \texttt{<SEG>} representation while
maintaining medical reasoning ability.

\subsection{Reasoning Segmentation and SAM Integration}
LISA~\cite{lai2024lisa} pioneered reasoning segmentation by
introducing a \texttt{<SEG>} token that bridges a VLM and SAM in the
general domain.
GLaMM~\cite{rasheed2024glamm} further enables grounded conversation
with pixel-level masks.
Subsequent work extends this line with finer-grained pixel reasoning
and generalized referring segmentation, including
PixelLM~\cite{ren2024pixellm} and Osprey~\cite{yuan2024osprey},
while open-vocabulary segmenters such as
SEEM~\cite{zou2023seem} and grounding models like
Grounding~DINO~\cite{liu2024groundingdino} broaden promptable
detection and segmentation.
In the medical domain, these ideas have been adapted with additional
task-specific modules; DuSSS~\cite{pan2025dusss} uses semantic
vision--language supervision for medical segmentation, and
CG-Reasoner~\cite{polamreddy2026cgreasoner} studies positional
reasoning segmentation in medical imaging. Foundation segmenters such as
MedSAM~\cite{ma2024segment} adapt SAM to medical imagery via
large-scale fine-tuning but lack language-level reasoning.
The key difference in our work is the training regime: we operate under
full-parameter optimization and use a staged curriculum to manage the
representational interference introduced by the segmentation interface.

\subsection{Multi-task Optimization}
Multi-task learning methods such as GradNorm~\cite{chen2018gradnorm},
PCGrad and gradient surgery~\cite{yu2020gradient}, and
Nash-MTL~\cite{navon2022nashmtl}
address competing objectives in general settings by re-weighting,
projecting, or bargaining over gradients, building on classical
multi-objective formulations~\cite{sener2018mgda} and uncertainty-based
loss weighting~\cite{kendall2018multitask}.
Our approach is more specific to SAM--VLM integration: rather than applying
a generic operation to all shared parameters, we use the hidden-state
distribution of the \texttt{<SEG>} token as an interpretable signal for the
segmentation interface and combine it with a staged curriculum.

\section{Methodology}

This section presents the overall framework.
To address this issue, we treat the \texttt{<SEG>} hidden state as the
semantic-to-spatial bridge that tells the mask decoder what to segment, and
we design the training process to make this bridge separable and stable.
We first describe the unified architecture, then analyze training
interference through the \texttt{<SEG>} hidden states, and finally detail
the parameter-decoupled training strategy.

\subsection{Unified Architecture}

\begin{figure*}[t]
  \centering
  \includegraphics[width=0.95\textwidth]{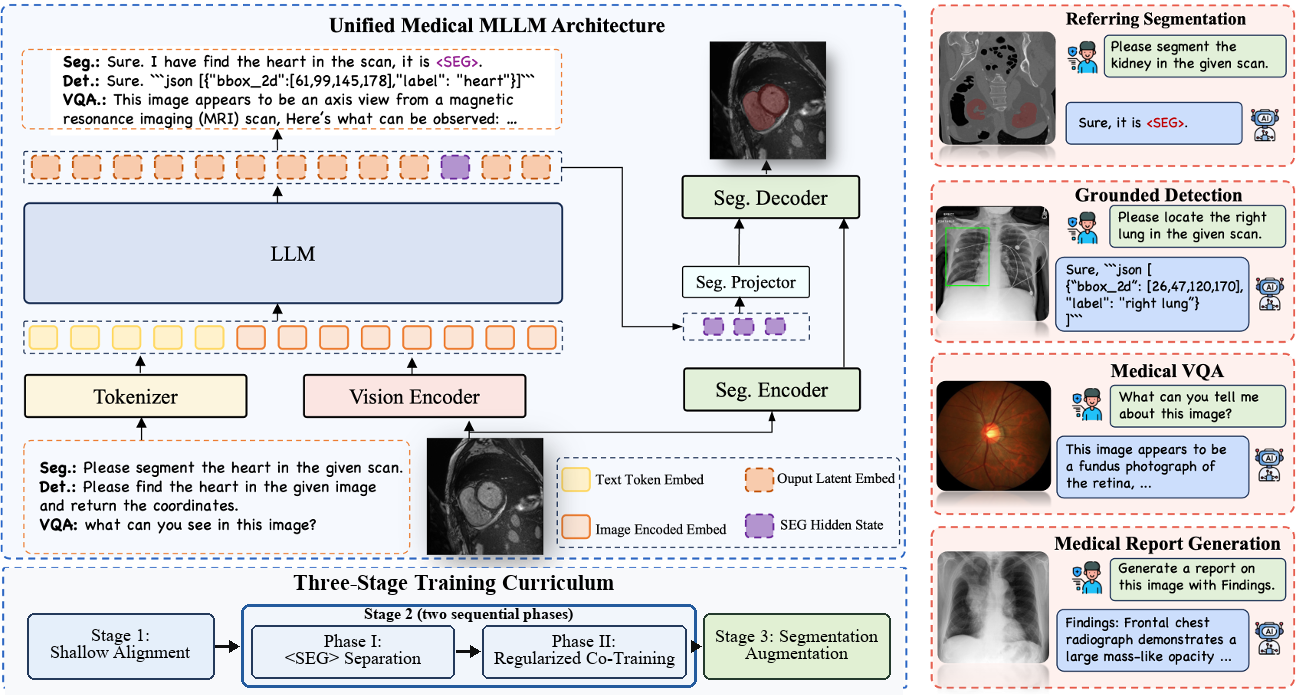}
  \caption{%
    Unified medical MLLM architecture and parameter-decoupled training overview.
    A VLM produces language responses for reasoning tasks and emits a
    \texttt{<SEG>} token for language-conditioned referring segmentation.
    The \texttt{<SEG>} hidden state is projected as an implicit prompt to
    the SAM2-based segmentation branch, while the bottom row summarizes the
    three training stages.}
  \label{fig:model_architecture}
\end{figure*}

Figure \ref{fig:model_architecture} shows the unified medical MLLM architecture designed in this work. Our model couples a multimodal VLM (Qwen2.5-VL~\cite{yang2025qwen2.5})
with a SAM2-based~\cite{ravi2024sam} segmentation head through a
lightweight projection module.
Given a medical image $\mathbf{x}$ and a textual instruction
$\mathbf{t}$, the VLM processes the input and produces a token
sequence.
When the instruction requests segmentation, the LLM generates a
special \texttt{<SEG>} token; its final-layer hidden state
$\mathbf{h}^{(\text{seg})} \in \mathbb{R}^d$ is projected to form an
implicit prompt that is fed into the SAM2 decoder to produce a
segmentation mask.
All other outputs (VQA answers, diagnostic text) are generated via
the standard language head.

\subsection{Training Interference and Hidden-State Dynamics}
\label{sec:training_interference}

Naive joint optimization exposes two forms of interference.
First, dense segmentation losses update the segmentation interface at a much
different scale and frequency from language-generation losses.
Second, the \texttt{<SEG>} token must simultaneously remain compatible with
the LLM hidden space and become an informative spatial prompt for SAM.
If its hidden states are entangled with generic linguistic tokens, the SAM
decoder receives ambiguous prompts and tends to produce unstable masks.

We therefore monitor the separability of \texttt{<SEG>} hidden states during
training.
For each validation segmentation sample, we collect the final-layer hidden
state of the generated \texttt{<SEG>} token and group samples by their target
anatomy or task label.
The Davies--Bouldin Index (DBI) provides a compact validation-time measure
of this structure: lower DBI indicates more compact within-class states and
larger between-class separation.
DBI is used only for monitoring and phase selection; it is not optimized by
backpropagation.

\begin{figure}[t]
  \centering
  \includegraphics[width=0.95\columnwidth]{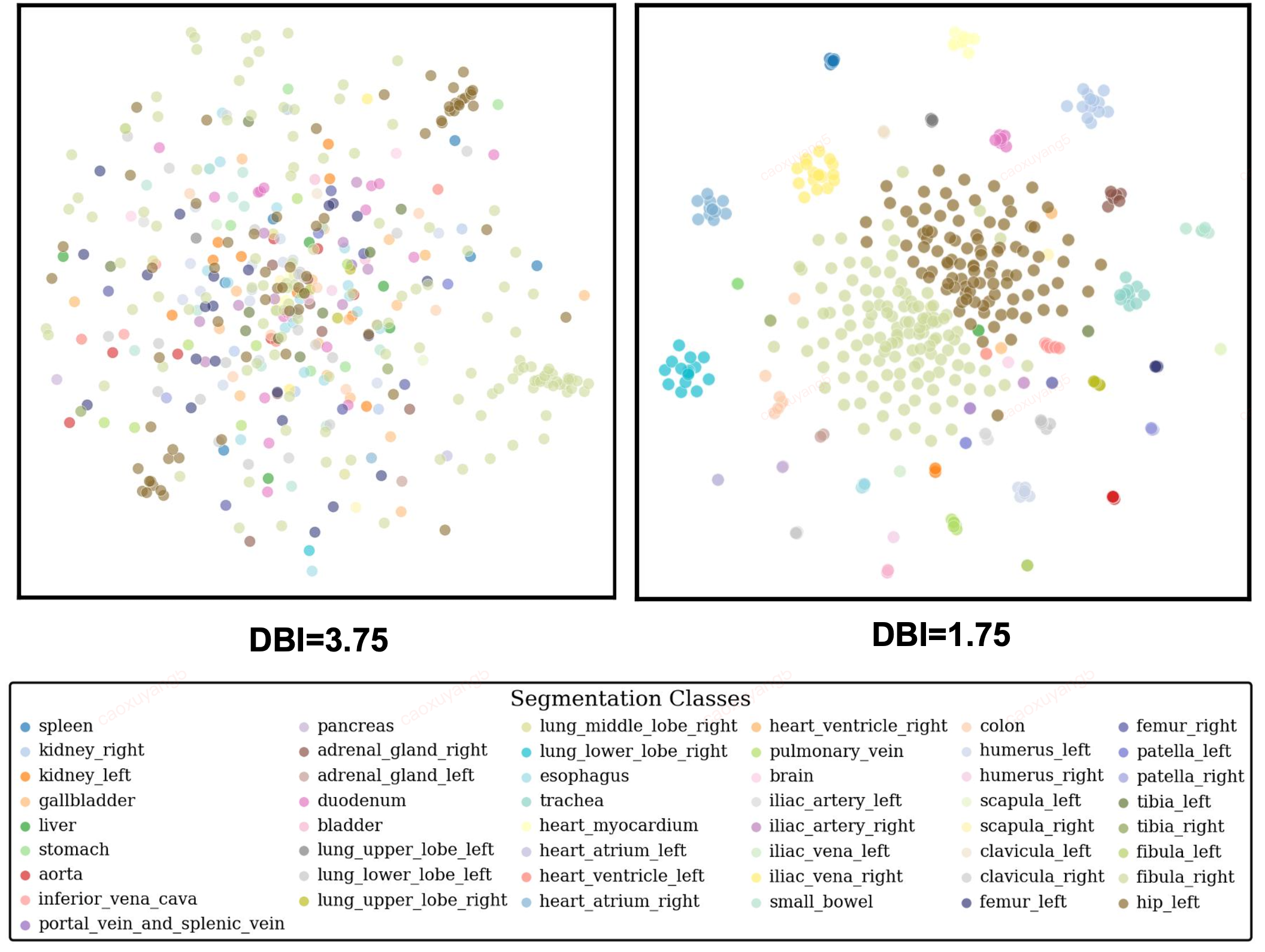}
  \caption{%
    t-SNE~\cite{van2008visualizing} visualization of \texttt{<SEG>} token
    hidden states as training progresses.
    DBI decreases from 3.75 to 1.75 across the two displayed
    checkpoints, showing that the implicit segmentation prompts become more
    compact and separable across target classes.}
  \label{fig:hidden_states_dynamic}
\end{figure}

Figure~\ref{fig:hidden_states_dynamic} visualizes the evolution of the
\texttt{<SEG>} state space.
As training proceeds, the hidden states become more separable and DBI drops
from 3.75 to 1.75 in the two displayed checkpoints, indicating that the
segmentation prompt interface becomes more structured before stronger joint
reasoning--segmentation optimization.

\subsection{Parameter-Decoupled Training Framework}
\label{sec:method}

The training process comprises three stages designed to progressively
build representational structure before introducing joint optimization, as
shown in Figure~\ref{fig:model_architecture}.
Algorithm~\ref{alg:training} specifies the training schedule,
phase-transition criterion, and gradient-scaling rule used in this process.

\begin{algorithm}[t]
\caption{Parameter-Decoupled Training}
\label{alg:training}
\textbf{Input}: Training data $\mathcal{D}$, validation set
$\mathcal{D}_\text{val}$, DBI threshold $\tau{=}1.9$,
gradient scale \revmath{\gamma_{\mathrm{seg}}{=}0.01}\\
\textbf{Output}: Trained unified model
\begin{algorithmic}[1]
\STATE \textbf{Stage 1 (Shallow Alignment):}
       Train vision encoder and projector only on image--caption
       pairs; LLM and SAM frozen.
\STATE \textbf{Stage 2, Phase I (Structured State Separation):}
       Train on segmentation instructions with the standard task loss;
       collect \texttt{<SEG>} hidden states for validation-time DBI monitoring.
\WHILE{$\mathrm{DBI}(\mathcal{D}_\text{val}) \geq \tau$}
    \STATE Update for one epoch.
    \STATE Compute DBI on $\mathcal{D}_\text{val}$ without backpropagation.
\ENDWHILE
\STATE \textbf{Stage 2, Phase II (Regularized Co-Training):}
       Scale segmentation gradients flowing from the \texttt{<SEG>}
       interface into the LLM backbone by \revmath{\gamma_{\mathrm{seg}}}.
       Train with full joint loss $\mathcal{L}_{\text{II}}$.
\STATE \textbf{Stage 3 (Segmentation Augmentation):}
       Freeze MLLM; fine-tune SAM only on segmentation data.
\end{algorithmic}
\end{algorithm}

\subsubsection{Stage 1: Medical Shallow Alignment.}
This stage is motivated by the domain gap between generic visual pretraining and
medical image interpretation.
Medical images differ substantially from natural images in texture, contrast,
viewpoint, and modality-specific artifacts, so directly mixing dense
segmentation losses with instruction tuning can force the LLM to compensate
for an under-aligned visual representation.
Before exposing the shared backbone to mask supervision, we therefore adapt
the visual front-end to medical image statistics using abundant image--caption
pairs.
Specifically, most parameters are frozen, and only the vision encoder and
projector are updated.
This provides low-risk medical visual--textual grounding: the model learns
modality and anatomy cues while the language backbone and SAM decoder remain
protected from noisy or sparse mask gradients.
This creates a stable semantic basis for the later \texttt{<SEG>} interface,
where segmentation supervision must refine spatial grounding rather than
repair basic medical visual recognition.

\subsubsection{Stage 2, Phase I: Structured State Separation.}
Phase~I is designed to let the \texttt{<SEG>} representation become
separable before stronger joint co-training is introduced.
Importantly, we do \emph{not} optimize a clustering loss, nor do we include
DBI in the training objective.
Instead, the model is updated with the standard segmentation-instruction task
loss, and the hidden-state geometry is measured on a held-out validation split
after each epoch.
For each validation sample $i$, let
$\mathbf{h}^{(\text{seg})}_i \in \mathbb{R}^d$ denote the final-layer hidden
state of the generated \texttt{<SEG>} token, with anatomy/task class label
$y_i \in \{1,\ldots,K\}$.
We group these states by label and compute per-class centroids only for
measuring separation, not for gradient-based optimization.

\subsubsection{DBI-Guided Phase Transition.}
After each epoch in Phase~I, we evaluate the Davies--Bouldin
Index~\cite{davies2009cluster} on a held-out validation split:
\begin{equation}
  \mathrm{DBI}
  = \frac{1}{K}\sum_{i=1}^{K}
    \max_{j \neq i}
    \frac{\sigma_i + \sigma_j}{d(c_i,\,c_j)},
  \label{eq:dbi}
\end{equation}
where $\sigma_k$ is the average intra-cluster distance for cluster $k$
and $d(c_i,c_j)$ is the Euclidean distance between centroids $c_i$
and $c_j$.
We transition from Phase~I to Phase~II when $\mathrm{DBI} < 1.9$ on the
held-out validation split.
This empirical threshold corresponds to a visibly separated \texttt{<SEG>}
state space in our runs and is typically reached after 2--4 epochs.
Importantly, DBI is monitoring-only and does not enter backpropagation.

\subsubsection{Stage 2, Phase II: Regularized Co-Training.}
In Phase~II, all MLLM parameters are optimized jointly.
To reduce interference with language representations, we apply
module-specific gradient scaling to segmentation-related updates.
Gradients flowing from the \texttt{<SEG>} projection interface into the LLM
backbone are multiplied by \revmath{\gamma_{\mathrm{seg}} = 0.01}, while language-modeling losses
and the SAM encoder/decoder updates are left unscaled.
The combined loss is:
\begin{equation}
  \begin{aligned}
  \mathcal{L}_{\text{II}}
  &= \lambda_1 \mathcal{L}_{\text{per-token}}
  + \lambda_2 \mathcal{L}_{\text{Dice}}
  + \lambda_3 \mathcal{L}_{\text{BCE}}, \\
  &\quad \lambda_1{=}1,\ \lambda_2{=}2,\ \lambda_3{=}1.
  \end{aligned}
  \label{eq:loss_phaseII}
\end{equation}

\subsubsection{Stage 3: Segmentation Augmentation.}
Stage~3 targets a different bottleneck from Stage~2.
Once the \texttt{<SEG>} token has learned to carry a stable semantic prompt,
the remaining segmentation errors are often caused by limited mask-decoder
adaptation to medical image appearance, including modality-specific texture,
low-contrast boundaries, and small structures.
Continuing to update the language backbone at this point may improve masks
slightly, but risks erasing the reasoning and instruction-following behavior
established during co-training.
Therefore, the entire MLLM is frozen, and only the SAM2-based segmentation
module is fine-tuned on additional segmentation data.
This keeps the semantic interface fixed while specializing the mask decoder
to medical masks, improving boundary precision without disturbing the
MLLM's multimodal reasoning capacity.
No inference overhead is introduced: at test time, the model performs
a single forward pass.

\section{Experiments and Results}

\begin{table*}[t]
  \centering
  \caption{%
    Results on multi-modal referring segmentation (Dice, \%) and
    medical detection (precision@0.5).
    Bold: best; underline: second best; --: not supported;
    $^\star$: zero-shot; $^\dagger$: reproduced/evaluation-only
    baseline.}
  \label{tab:results_medical_segmentation_detection}
  \setlength{\tabcolsep}{2.5pt}
  \fontsize{9}{10}\selectfont
  \begin{tabular}{lcccccccc|c}
    \toprule
    \multirow{2}{*}{\textbf{Model}} &
    \multicolumn{8}{c}{\textbf{MeCOVQA-G+}$_{seg}$} &
    \textbf{MedSAM2}$_{det}$ \\
    & \textbf{DER} & \textbf{CT} & \textbf{PET} & \textbf{X-RAY}
    & \textbf{END} & \textbf{MR} & \textbf{US} & \textbf{FP} & \\
    \midrule
    MedPLIB 14B~\cite{huang2024towards}
      & 79.84 & 57.58 & 64.25 & 8.47$^\star$  & 44.35$^\star$
      & 27.38$^\star$ & 34.22$^\star$ & 4.82$^\star$ & -- \\
    UniBioMed$^\dagger$~\cite{wu2025unibiomed}
      & 63.29 & 14.07 & 19.03 & \textbf{29.86}
      & 67.35 & 25.74 & 0.18 & 69.22 & -- \\
    Qwen2.5-VL-7B+SAM2$^\dagger$~\cite{yang2025qwen2.5,ravi2024sam}
      & 32.97 & 19.31 & 13.98 & 5.75
      & 52.91 & 7.79 & 8.48 & 0.32 & 20.90 \\
    Qwen2.5-VL-32B+SAM2$^\dagger$~\cite{yang2025qwen2.5,ravi2024sam}
      & 34.94 & 23.23 & 22.84 & 4.81
      & 58.55 & 10.75 & 10.55 & 0.22 & 30.10 \\
    \midrule
    \textbf{Ours-8B}
      & \textbf{92.09} & \underline{64.04} & \underline{77.93} & 14.69
      & \underline{92.80} & \underline{43.07} & \underline{83.83} & \underline{74.07} & \underline{44.60} \\
    \textbf{Ours-33B}
      & \underline{91.45} & \textbf{73.87} & \textbf{79.25} & \underline{29.20}
      & \textbf{93.10} & \textbf{48.19} & \textbf{86.10} & \textbf{96.02} & \textbf{44.90} \\
    \bottomrule
  \end{tabular}
\end{table*}

\begin{table*}[t]
  \centering
  \caption{%
    Segmentation results on MedSegBench~\cite{kucs2024medsegbench}
    (Dice, \%).
    Bold: best; underline: second best; --: not reported.}
  \label{tab:results_medsegbench}
  \setlength{\tabcolsep}{4pt}
  \resizebox{\textwidth}{!}{%
  \begin{tabular}{lcccccccc}
    \toprule
    \textbf{Model} &
    \textbf{ISIC16} & \textbf{Kvasir} & \textbf{IDRiD} &
    \textbf{CovidQUEx} & \textbf{Promise12} & \textbf{MosMed+} &
    \textbf{US-Nerve} & \textbf{TNBC} \\
    \midrule
    Unet-MIT         & 89.10 & 56.90 & 5.30  & --    & --    & 76.10          & --    & 75.90 \\
    Unet-EfficientNet & 90.30 & 81.20 & 7.80  & 74.40 & 89.20 & 78.10          & 78.70 & 73.80 \\
    Unet-MobileNetV2 & 89.10 & 75.40 & 9.20  & 74.20 & 89.60 & 78.50          & 77.20 & 76.20 \\
    Unet-DenseNet121 & 89.30 & 79.40 & 8.90  & 75.60 & 90.00 & \textbf{79.10} & 78.60 & 78.80 \\
    Unet-ResNet50    & 88.70 & 69.80 & 9.00  & 73.40 & 88.80 & \underline{79.00} & 77.60 & 78.50 \\
    \midrule
    \textbf{Ours-8B}  & \underline{93.16} & \underline{90.91} & \underline{47.10} & \underline{77.84}
                      & \underline{90.81} & 78.17             & \underline{80.15} & \textbf{82.93} \\
    \textbf{Ours-33B} & \textbf{93.71}    & \textbf{91.22}    & \textbf{56.86}    & \textbf{78.39}
                      & \textbf{91.40}    & 78.01             & \textbf{80.24}    & \underline{81.20} \\
    \bottomrule
  \end{tabular}
  }
\end{table*}

\subsection{Implementation Details}

We train on $4{\times}8$ NVIDIA H200 GPUs with a global batch size of
256, using AdamW (weight decay $0.01$), with learning rates $1{\times}
10^{-5}$ for the LLM, $2{\times}10^{-6}$ for the vision transformer,
and $1{\times}10^{-5}$ for the alignment module.
A cosine scheduler is applied.
DeepSpeed~\cite{rasley2020deepspeed} and
FlashAttention-2~\cite{dao2023flashattention} are used for memory
efficiency.
All models are evaluated with the same prompt templates and decoding settings
within each benchmark.
For language tasks, we report accuracy after answer normalization.
For referring segmentation, we report Dice score.
For detection, we report precision at IoU threshold 0.5.

\paragraph{Computational Cost.}
Our three-stage training requires approximately 3 days for Ours-8B and
7 days for Ours-33B on $4{\times}8$ NVIDIA H200 GPUs.
The staged procedure increases training cost compared with simpler joint
fine-tuning, but it does not introduce additional inference overhead: at
test time, all tasks use the same forward pass and the same segmentation
branch.

\paragraph{Model Variants}


We construct two model variants with different parameter scales: Ours-8B and Ours-33B. Ours-8B is built upon Qwen2.5-VL-7B~\cite{yang2025qwen2.5}, which consists of a 7B-parameter LLM and a ViT-based vision encoder, while Ours-33B adopts Qwen2.5-VL-32B with a 32B-parameter LLM. All remaining components—including the SAM2-based segmentation head, the \texttt{<SEG>} projection layer, training hyperparameters, and the three-stage training pipeline—are kept identical across the two variants.

\subsection{Datasets and Benchmarks}

\subsubsection{Training Data.}
The training corpus combines public and synthesized data for medical image
captioning, report generation, VQA, textual QA, general instruction following,
and language-conditioned segmentation. These sources support different
curriculum stages: image--caption pairs provide shallow medical alignment,
instruction data preserves reasoning and dialogue ability, and segmentation
samples teach textual queries to ground into masks.
After quality control and deduplication, the final set contains 16.8M
public samples and 1.6M synthesized samples ($>3$B text tokens,
12.6M images).
See Appendix~A for public training datasets.

\subsubsection{Evaluation Benchmarks.}
We evaluate the model along three axes.
\textit{Medical textual QA} tests knowledge preservation after segmentation
training on \revtext{five public benchmarks}:
PubMedQA~\cite{jin2019pubmedqa}, MedMCQA~\cite{pal2022medmcqa},
MedQA~\cite{jin2021disease}, MedXpertQA~\cite{zuo2025medxpertqa},
and CMMLU~\cite{li2024cmmlu}.
\textit{Medical visual QA} evaluates image-level clinical understanding:
VQA-RAD~\cite{lau2018dataset},
MedXpertQA~\cite{zuo2025medxpertqa}, SLAKE~\cite{liu2021slake},
PATH-VQA~\cite{naseem2022vision}, and PMC-VQA~\cite{zhang2023pmc}.
\textit{Medical referring segmentation and detection} evaluates pixel-level
grounding: MedSegBench~\cite{kucs2024medsegbench} (Dice),
MeCOVQA-G$+_{seg}$~\cite{wang2025citrus} (Dice), and
MedSAM2$_{det}$~\cite{wang2025citrus} (precision@0.5).
See Appendix~B for public evaluation benchmarks.

\subsection{Results}

We first report pixel-level referring segmentation and grounding results,
then evaluate whether the same model retains image-level and textual medical
reasoning ability.
This ordering reflects the central question of our study: whether adding a
SAM-style segmentation branch through a \texttt{<SEG>} interface can improve
localization without collapsing the VLM's reasoning capability.

\subsubsection{Results on Segmentation Benchmarks.}

Table~\ref{tab:results_medical_segmentation_detection} first evaluates the
language-conditioned grounding setting most aligned with our goal: translating
a clinical expression into a \texttt{<SEG>} prompt for mask prediction.
To make the tool-composed baseline explicit, we reproduce
Qwen2.5-VL+SAM2 in an evaluation-only manner: Qwen2.5-VL is prompted to
return a JSON bounding box for the queried region, the
parsed box is used as the spatial prompt to SAM2, and invalid or missing boxes
are counted as grounding failures. This tests whether a strong VLM can provide
sufficient spatial prompts for an off-the-shelf segmenter without learning a
dedicated segmentation interface. The reproduced 7B/32B pipelines
obtain only 32.97/34.94 DER Dice and 20.90/30.10 precision on
MedSAM2$_{det}$, indicating that loosely attaching SAM2 to a strong VLM is not
enough for robust medical referring segmentation.

Table~\ref{tab:results_medsegbench} then compares against the
dataset-specific expert segmenters used in MedSegBench.
Unlike these official U-Net baselines, trained \emph{per-dataset} with one specialized model for each
benchmark~\cite{kucs2024medsegbench}, our model is trained jointly and uses a
single language-conditioned interface across datasets.
These expert baselines instantiate U-Net~\cite{ronneberger2015u} with common
vision backbones including ResNet~\cite{he2016deep},
EfficientNet~\cite{tan2019efficientnet}, MobileNetV2~\cite{sandler2018mobilenetv2},
and DenseNet~\cite{iandola2014densenet}.
Despite this one-model-versus-many-experts setting, Ours-33B surpasses
the dataset-specialized experts on most MedSegBench datasets and remains
competitive on the rest, complementing Table~\ref{tab:results_medical_segmentation_detection}.
Quantitatively, Ours-33B ranks first on six of eight datasets, with especially
large gains on IDRiD (56.86 vs. 9.20) and Kvasir (91.22 vs. 81.20).

\begin{table*}[t]
  \centering
  \caption{%
    Medical visual question answering results (accuracy, \%).
    Bold: best; underline: second best.}
  \label{tab:results_medical_vqa}
  \setlength{\tabcolsep}{5pt}
  \resizebox{\textwidth}{!}{%
  \begin{tabular}{lcccccc}
    \toprule
    \textbf{Model} & \textbf{VQA-RAD} & \textbf{MedXpertQA}
    & \textbf{SLAKE} & \textbf{PATH-VQA} & \textbf{PMC-VQA} & \textbf{Avg.} \\
    \midrule
    Qwen2.5-VL 7B~\cite{yang2025qwen2.5}   & 66.30 & 20.75 & 67.86 & 42.30 & 50.86 & 49.61 \\
    Lingshu 7B~\cite{xu2025lingshu}         & 68.74 & 26.90 & 82.90 & 60.23 & 55.77 & 58.91 \\
    HealthGPT 14B~\cite{lin2025healthgpt}   & 64.08 & 24.55 & 67.43 & 58.67 & 56.90 & 54.33 \\
    MedGemma 27B~\cite{sellergren2025medgemma} & 63.86 & \underline{33.10} & 76.17 & 47.60 & 45.35 & 53.22 \\
    Qwen2.5-VL 32B~\cite{yang2025qwen2.5}  & 72.28 & 25.30 & 76.36 & 41.58 & 53.58 & 53.82 \\
    Lingshu 32B~\cite{xu2025lingshu}        & \underline{75.39} & 31.00 & \underline{87.68} & \textbf{64.76} & \underline{57.23} & \underline{63.21} \\
    HealthGPT 32B~\cite{lin2025healthgpt}   & 64.75 & 26.40 & 70.58 & 62.93 & 54.93 & 55.92 \\
    HuatuoGPT-V 34B~\cite{chen2024towards}  & 63.64 & 22.65 & 73.02 & 44.92 & 56.79 & 52.20 \\
    GPT-5                                   & 68.37 & \textbf{51.48} & 65.82 & 31.74 & 36.10 & 50.70 \\
    \midrule
    \textbf{Ours-8B}  & 64.30 & 25.10 & 84.91 & 62.00 & 55.64 & 58.39 \\
    \textbf{Ours-33B} & \textbf{77.83} & 29.15 & \textbf{88.40} & \underline{63.89} & \textbf{59.74} & \textbf{63.80} \\
    \bottomrule
  \end{tabular}
  }
\end{table*}

\begin{table*}[t]
  \centering
  \caption{%
    Medical textual question answering results (accuracy, \%).
    Bold: best; underline: second best.}
  \label{tab:results_medical_text_qa}
  \setlength{\tabcolsep}{5pt}
  \resizebox{\textwidth}{!}{%
  \begin{tabular}{lcccccc}
    \toprule
    \textbf{Model} & \textbf{PubMedQA} & \textbf{MedMCQA}
    & \textbf{MedQA} & \textbf{MedXpertQA} & \textbf{CMMLU} & \textbf{Avg.} \\
    \midrule
    Qwen2.5-VL 7B~\cite{yang2025qwen2.5}   & 75.80 & 53.40 & 57.50 & 12.40 & 68.80 & 53.58 \\
    Lingshu 7B~\cite{xu2025lingshu}         & 75.40 & 56.13 & 63.39 & 16.45 & 69.02 & 56.08 \\
    HealthGPT 14B~\cite{lin2025healthgpt}   & 69.40 & 63.33 & 66.93 & 12.45 & 55.36 & 53.49 \\
    MedGemma 27B~\cite{sellergren2025medgemma} & \textbf{79.00} & 63.23 & \textbf{81.15} & 22.01 & 60.24 & \revcell{61.13} \\
    Qwen2.5-VL 32B~\cite{yang2025qwen2.5}  & 68.60 & 62.71 & 71.33 & 15.88 & 82.60 & 60.22 \\
    Lingshu 32B~\cite{xu2025lingshu}        & 78.20 & \underline{65.05} & 74.94 & \underline{22.86} & 82.37 & 64.69 \\
    HealthGPT 32B~\cite{lin2025healthgpt}   & 74.20 & 64.04 & 68.89 & 13.84 & 69.47 & 58.09 \\
    HuatuoGPT-V 34B~\cite{chen2024towards}  & 71.00 & 55.08 & 58.52 & 12.20 & 77.64 & 54.89 \\
    GPT-5                                   & 78.00 & 62.99 & 76.96 & \textbf{40.75} & \underline{82.93} & \textbf{68.33} \\
    \midrule
    \textbf{Ours-8B}  & 74.80 & 55.10 & 64.89 & 16.90 & 71.19 & 56.58 \\
    \textbf{Ours-33B} & \underline{78.40} & \textbf{65.62} & \underline{80.28} & 22.20 & \textbf{83.27} & \underline{65.95} \\
    \bottomrule
  \end{tabular}
  }
\end{table*}

\begin{table}[!ht]
  \centering
  \caption{Ablation on two-phase instruction fine-tuning (Ours-8B).}
  \label{tab:ablation_two_phase}
  \setlength{\tabcolsep}{4pt}
  \begin{tabular}{lccc}
    \toprule
    \textbf{Model} & \textbf{VQA} & \textbf{Text QA} & \textbf{Dice} \\
    \midrule
    w/o two-phase & 56.67 & 49.27 & 64.92 \\
    w/ two-phase  & \textbf{58.39} & \textbf{56.58} & \textbf{80.13} \\
    \bottomrule
  \end{tabular}
\end{table}

\begin{figure*}[t]
  \centering
  \includegraphics[width=0.98\textwidth]{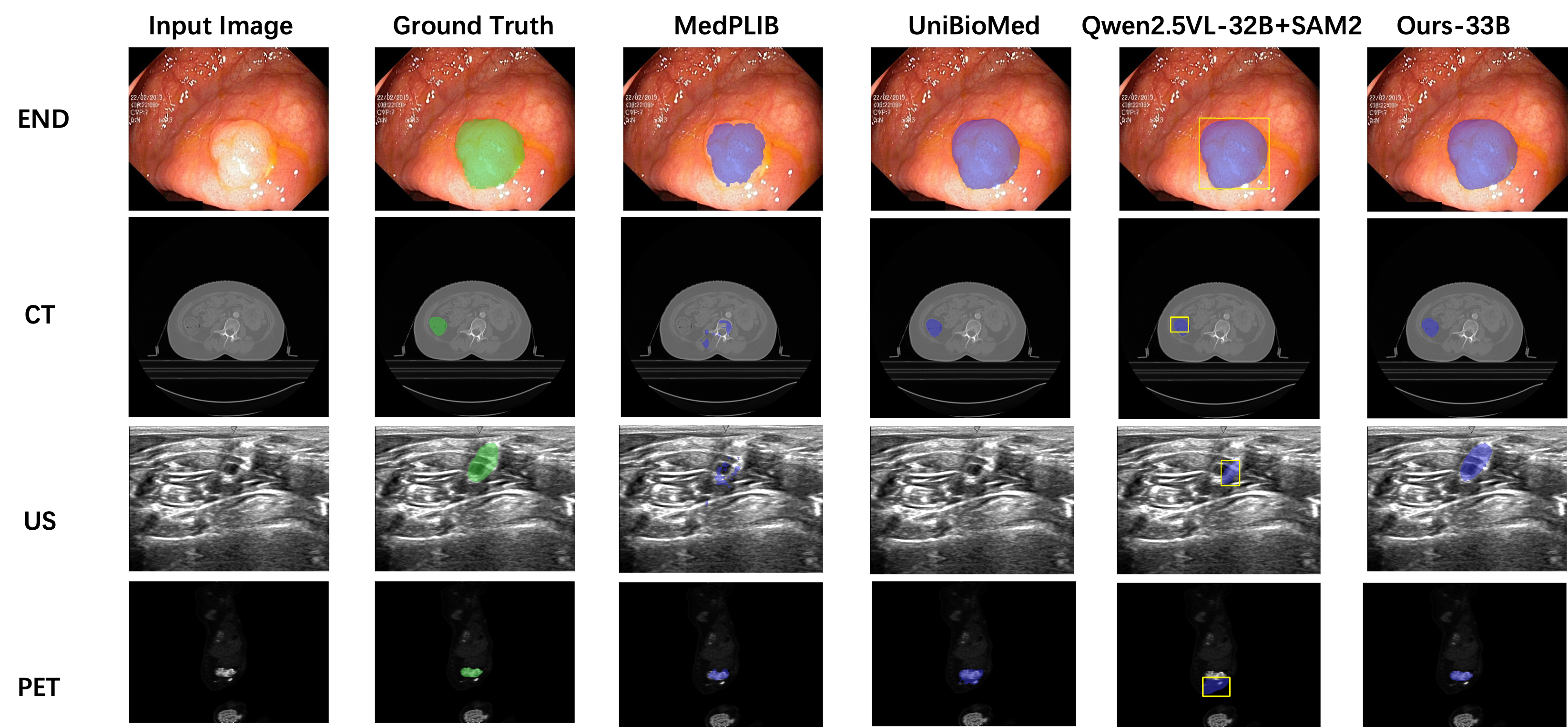}
  \caption{%
    Qualitative comparison of medical referring segmentation results.
    We compare MedPLIB, UniBioMed, Qwen2.5-VL-32B+SAM2, and Ours-33B across
    representative imaging modalities.
    Green masks denote ground truth, all predicted segmentation masks are shown
    in blue, and yellow boxes indicate the grounding output used by the
    Qwen2.5-VL+SAM2 pipeline.}
  \label{fig:qualitative_results}
\end{figure*}

\paragraph{Qualitative Analysis.}
Figure~\ref{fig:qualitative_results} visualizes differences among
MedPLIB, UniBioMed, Qwen2.5-VL-32B+SAM2, and Ours-33B, highlighting failure
modes such as coarse language-to-box grounding and boundary leakage.
The Qwen2.5-VL+SAM2 pipeline often localizes only a rough box before mask
generation, while MedPLIB and UniBioMed can miss small structures or leak into
nearby tissues. Ours-33B produces masks that more closely follow the green
annotations across modalities, especially around low-contrast boundaries.

\subsubsection{Results on Medical VQA Benchmarks.}

\begin{table}[!ht]
  \centering
  \caption{Sensitivity to gradient scaling factor \revmath{\gamma_{\mathrm{seg}}} (Ours-8B).}
  \label{tab:hookgrad_professional}
  \setlength{\tabcolsep}{4pt}
  \begin{tabular}{lcccc}
    \toprule
    \textbf{Metric}
      & $1.0$ & $0.3$ & $0.01$ & $0.001$ \\
    \midrule
    VQA Avg.  & 56.23 & 56.90 & 57.93 & \textbf{58.84} \\
    Text Avg. & 54.22 & 54.92 & 55.60 & \textbf{56.13} \\
    Dice Avg. & \textbf{65.95} & 62.39 & 57.42 & 50.38 \\
    \bottomrule
  \end{tabular}
\end{table}

\begin{table*}[!ht]
  \centering
  \caption{Ablation on Stage~3 segmentation augmentation (Ours-33B, Dice, \%).}
  \label{tab:ablation_segmentation_augmentation}
  \setlength{\tabcolsep}{3pt}
  \fontsize{9}{10}\selectfont
  \begin{tabular}{lccccccccc}
    \toprule
    \textbf{Model} & \textbf{ISIC16} & \textbf{Kvasir} & \textbf{IDRiD}
    & \textbf{CovidQUEx} & \textbf{Promise12} & \textbf{MosMed+}
    & \textbf{US-Nerve} & \textbf{TNBC} & \textbf{Avg.} \\
    \midrule
    w/o Stage~3 & 86.19 & 83.10 & 41.34 & 72.49 & 89.99 & \textbf{79.68} & 80.10 & 77.01 & 76.24 \\
    w/ Stage~3  & \textbf{93.71} & \textbf{91.22} & \textbf{56.86} & \textbf{78.39}
                & \textbf{91.40} & 78.01 & \textbf{80.24} & \textbf{81.20} & \textbf{81.38} \\
    \bottomrule
  \end{tabular}
\end{table*}

Ours-33B achieves the highest average accuracy (63.80\%) among all
evaluated methods, establishing new best results on VQA-RAD, SLAKE, and
PMC-VQA.
This is important because the model is not optimized solely for VQA; it is
also trained to emit \texttt{<SEG>} tokens and support segmentation outputs.
The result suggests that the proposed parameter-decoupled curriculum improves
pixel-level capability while preserving image-level clinical semantics and
question answering ability.
The gains are not uniform across benchmarks: Ours-33B is strongest on VQA-RAD,
SLAKE, and PMC-VQA, but remains below GPT-5 on MedXpertQA.
This indicates that the model is competitive on image-grounded medical VQA,
while harder expert-level reasoning still benefits from larger general-purpose
language capacity.

\subsubsection{Results on Medical Textual QA Benchmarks.}

Ours-33B achieves second-best average accuracy (65.95\%), best on MedMCQA
and CMMLU, and top-two results on PubMedQA and MedQA.
These benchmarks do not require image segmentation, so they provide a useful
stress test for catastrophic forgetting of medical knowledge during
segmentation-oriented training.
The strong textual QA performance indicates that freezing and scaling choices
in the training curriculum help preserve the language backbone's medical
reasoning ability while enabling pixel-level segmentation.
This table is especially important because none of these benchmarks requires
mask prediction; improvements here cannot be explained by the SAM2 branch.
Instead, they indicate that the staged training procedure preserves text-only
medical knowledge while improving segmentation, avoiding the common trade-off
between localization and language reasoning.

\subsubsection{Ablation Study.}

We ablate the key components of the proposed curriculum to verify their
separate effects on reasoning and segmentation.
The two-phase comparison evaluates controlled segmentation exposure, the
gradient-scaling sweep examines language-backbone updates, and the Stage~3
ablation isolates VLM-frozen SAM2 specialization.

Table~\ref{tab:ablation_two_phase} shows that two-phase instruction
fine-tuning substantially improves both textual QA and segmentation,
indicating that controlled exposure to segmentation supervision is important
for maintaining the unified model.
Without this separation, segmentation supervision is introduced too abruptly
and both the \texttt{<SEG>} prompt space and the language reasoning ability
are less stable.

Table~\ref{tab:hookgrad_professional} reveals the expected trade-off of
\revmath{\gamma_{\mathrm{seg}}}: smaller values better preserve QA performance but weaken the
segmentation signal.
We use \revmath{\gamma_{\mathrm{seg}}{=}0.01} as a practical compromise before the final
segmentation specialization stage, balancing reasoning retention and prompt
learning across tasks.

Table~\ref{tab:ablation_segmentation_augmentation} shows that
Stage~3 improves the Ours-33B average MedSegBench Dice from 76.24\% to
81.38\% while keeping the VLM frozen, consistent with the goal of improving
masks without altering the reasoning backbone or instruction-following
behavior learned earlier.

\section{Conclusion}

We present a parameter-decoupled training framework for unified medical
reasoning and language-conditioned segmentation.
Rather than framing SAM--VLM integration as only an architectural problem,
we focus on how the \texttt{<SEG>} interface should be trained under
full-parameter optimization.
The proposed curriculum uses medical shallow alignment, DBI-monitored
instruction tuning with gradient scaling, and VLM-frozen segmentation
specialization to improve pixel-level masks while preserving medical VQA and
textual QA performance.
Extensive experiments on diverse benchmarks demonstrate accurate anatomical
delineation without sacrificing reasoning performance, establishing an
effective paradigm for unified medical multimodal modeling.

\clearpage
\appendix
\begin{center}
  {\LARGE\bfseries Appendix}
\end{center}
\vspace{1em}

\section{Public Training Datasets}

This appendix summarizes the public training datasets and curation procedure
used in our training corpus. Following our prior data construction protocol, we
aggregate open-source resources into five categories: medical image--caption,
medical multimodal instruction/VQA, referring segmentation/grounding, medical
textual QA and reasoning, and general instruction data.

\subsection{Dataset Summary}

Each dataset in Table~\ref{tab:appendix_training_datasets} is annotated with
its corresponding public reference.

\subsection{Data Processing and Quality Control}

We apply a unified data processing pipeline to textual, visual, and multimodal
samples before training. For image data, low-resolution images are removed using
a 4096-pixel threshold. For DICOM-based medical imaging data, we further check
physical dimension consistency, orientation consistency, and slice-spacing
consistency, and remove samples with corrupted or inconsistent metadata. Private
or patient-identifying information in metadata is removed during
de-identification.

For textual samples, we use regular-expression filters and token-length
thresholds to remove irrelevant or low-quality instructions, especially samples
that are too short or exceed the intended training length. Pure text
instruction--response pairs are deduplicated to reduce redundancy. For
image--text pairs, we combine text-level matching with image hashing to remove
duplicated multimodal samples. Conversational medical data are additionally
de-identified with LLM-based rewriting, retaining only clinically relevant
consultation content while removing personal information.

We also perform lightweight model-based enhancement for public VQA-style data
whose answers are overly brief or whose questions are highly repetitive. A VLM
is used to consolidate multiple simple questions about the same image into more
holistic instructions, improving coverage of image interpretation, feature
relations, and diagnostic reasoning.

For referring segmentation and grounding tasks, we use
SA-Med2D-20M~\cite{ye2023samed2d20m}. SA-Med2D-20M is a large-scale 2D
medical segmentation corpus built from public medical imaging datasets. It
provides paired images and dense masks across diverse modalities, anatomical
regions, and lesion types, making it suitable for broad medical grounding
supervision. For referring segmentation data, we use the image--mask pairs and
GPT-5 to construct instruction-style samples whose textual prompts refer to
target anatomical or pathological regions. For grounding data, we first convert
each segmentation mask into a tight bounding box and then use GPT-5 to build
dialogue-style localization samples. Finally, medical data are tagged by imaging
modality, task type, and anatomical region, enabling transparent corpus analysis
and stratified evaluation across downstream tasks.

\begin{table}[!htbp]
  \centering
  \caption{List of open-source datasets collected for training.}
  \label{tab:appendix_training_datasets}
  \setlength{\tabcolsep}{3pt}
  \fontsize{9}{10}\selectfont
  \begin{tabular}{>{\raggedright\arraybackslash}p{0.21\textwidth}>{\raggedright\arraybackslash}p{0.73\textwidth}}
    \toprule
    \textbf{Category} & \textbf{Included Public Datasets} \\
    \midrule
    Image--caption & PMCOA~\cite{lin2023pmcclipcontrastivelanguageimagepretraining},
    ROCO~\cite{ROCO}, LLaVA-Med~\cite{li2023llava},
    MedPix2.0~\cite{siragusa2025medpix20},
    CheXpert Plus~\cite{chambon2024chexpertplusaugmentinglarge},
    MIMIC-CXR~\cite{johnson2019mimiccxr}, ROCOv2~\cite{ROCOv2},
    Quilt-LLaVA~\cite{seyfioglu2024quiltllava},
    PubMedVision~\cite{chen2024huatuogptvision},
    IU-Xray~\cite{demner2016preparing-iu-xray} \\
    MM instruction & VQA-RAD~\cite{lau2018dataset},
    PMC-VQA~\cite{zhang2023pmc}, PATH-VQA~\cite{naseem2022vision},
    SLAKE~\cite{liu2021slake}, MIMIC-CXR-VQA~\cite{bae2024ehrxqa},
    VQA-Med-2019~\cite{ImageCLEFVQA-Med2019} \\
    Referring segmentation / grounding & SA-Med2D-20M~\cite{ye2023samed2d20m} \\
    Textual QA & JMed~\cite{wang2025citrusleveragingexpertcognitive},
    HealthCareMagic~\cite{li2023chatdoctormedicalchatmodel},
    iCliniq~\cite{li2023chatdoctormedicalchatmodel},
    HuatuoGPT2-SFT-GPT4~\cite{chen2023huatuogptii},
    Citrus-S3~\cite{wang2025citrusleveragingexpertcognitive},
    medical-o1-verifiable-problem~\cite{chen2024huatuogpto1medicalcomplexreasoning},
    Medical-R1-Distill-Data~\cite{chen2024huatuogpto1medicalcomplexreasoning},
    huatuogpt-o1-for-reasoning~\cite{chen2024huatuogpto1medicalcomplexreasoning},
    MedReason~\cite{wu2025medreasonelicitingfactualmedical},
    MedThoughts~\cite{medthoughts8k},
    medical-o1-reasoningSFT~\cite{chen2024huatuogpto1medicalcomplexreasoning},
    AlpaCare~\cite{zhang2025alpacareinstructiontunedlargelanguagemodels},
    ApolloCorpus~\cite{wang2024apollo}, MedQuAD~\cite{BenAbacha-BMC-2019},
    MedQA~\cite{jin2021disease},
    PMC-LLaMA~\cite{wu2023pmcllamabuildingopensourcelanguage} \\
    General instruction & LLaVA1.5~\cite{liu2024improvedbaselinesvisualinstruction},
    PixMo~\cite{deitke2024molmopixmoopenweights},
    ALLaVA~\cite{chen2024allava}, OpenHermes-2.5~\cite{OpenHermes2-5},
    OKVQA~\cite{marino2019okvqavisualquestionanswering},
    A-OKVQA~\cite{schwenk2022aokvqabenchmarkvisualquestion},
    OCRVQA~\cite{mishraICDAR19}, TextCaps~\cite{sidorov2019textcaps} \\
    \bottomrule
  \end{tabular}
\end{table}

\section{Public Evaluation Benchmarks}

Following the main paper, we organize evaluation benchmarks along three axes:
medical textual QA, medical visual QA, and medical referring
segmentation/detection. Table~\ref{tab:appendix_evaluation_benchmarks}
summarizes the public benchmarks, task types, modalities, and reported metrics.

\begin{table}[!htbp]
  \centering
  \caption{Public evaluation benchmarks used in the main experiments.}
  \label{tab:appendix_evaluation_benchmarks}
  \setlength{\tabcolsep}{3pt}
  \fontsize{9}{10}\selectfont
  \begin{tabular}{>{\raggedright\arraybackslash}p{0.16\textwidth}>{\raggedright\arraybackslash}p{0.50\textwidth}>{\raggedright\arraybackslash}p{0.14\textwidth}>{\raggedright\arraybackslash}p{0.12\textwidth}}
    \toprule
    \textbf{Axis} & \textbf{Benchmarks} & \textbf{Modality} & \textbf{Metric} \\
    \midrule
    Medical textual QA & \revcell{PubMedQA~\cite{jin2019pubmedqa},
    MedMCQA~\cite{pal2022medmcqa}, MedQA~\cite{jin2021disease},
    MedXpertQA~\cite{zuo2025medxpertqa}, CMMLU~\cite{li2024cmmlu}}
    & Text & Accuracy \\
    Medical visual QA & VQA-RAD~\cite{lau2018dataset},
    MedXpertQA~\cite{zuo2025medxpertqa}, SLAKE~\cite{liu2021slake},
    PATH-VQA~\cite{naseem2022vision}, PMC-VQA~\cite{zhang2023pmc}
    & Image+Text & Accuracy \\
    Referring segmentation & MedSegBench~\cite{kucs2024medsegbench},
    MeCOVQA-G$+_{seg}$~\cite{wang2025citrus}
    & Image+Text & Dice \\
    Grounding & MedSAM2$_{det}$~\cite{wang2025citrus}
    & Image+Text & Precision@0.5 \\
    \bottomrule
  \end{tabular}
\end{table}

MedSegBench~\cite{kucs2024medsegbench} further contains eight
dataset-specific segmentation subsets used in the main paper. These subsets
span dermatology, endoscopy, ophthalmology, radiography, MRI, CT, ultrasound,
and histopathology, allowing us to compare one unified model with multiple
expert segmenters trained separately for each dataset. Table~\ref{tab:appendix_medsegbench_subsets}
summarizes the domain and segmentation target of each subset.

\begin{table}[!htbp]
  \centering
  \caption{MedSegBench subsets used for segmentation evaluation.}
  \label{tab:appendix_medsegbench_subsets}
  \setlength{\tabcolsep}{4pt}
  \fontsize{9}{10}\selectfont
  \begin{tabular}{>{\raggedright\arraybackslash}p{0.18\textwidth}>{\raggedright\arraybackslash}p{0.18\textwidth}>{\raggedright\arraybackslash}p{0.45\textwidth}>{\raggedright\arraybackslash}p{0.09\textwidth}}
    \toprule
    \textbf{Subset} & \textbf{Domain / Modality} & \textbf{Segmentation Target} & \textbf{Metric} \\
    \midrule
    ISIC16~\cite{Isic2016MSBench} & Dermoscopy & Skin lesion region in dermoscopic images & Dice \\
    Kvasir~\cite{KvasirMSBench} & Endoscopy & Colorectal polyp region in colonoscopy images & Dice \\
    IDRiD~\cite{IdribMSBench} & Fundus photography & Retinal pathology or structure masks for diabetic retinopathy analysis & Dice \\
    CovidQUEx~\cite{CovidQUExMSBench} & Chest X-ray & Lung or infection-related regions in COVID-19 radiographs & Dice \\
    Promise12~\cite{Promise12MSBench} & Prostate MRI & Prostate gland region in pelvic MR images & Dice \\
    MosMed+~\cite{MosMedPlusMSBench} & Chest CT & Lung or infection-related regions in thoracic CT volumes & Dice \\
    US-Nerve~\cite{UltrasoundNerveMSBench} & Ultrasound & Peripheral nerve region in ultrasound images & Dice \\
    TNBC~\cite{TnbcnucleiMSBench} & Histopathology & Nuclei regions in triple-negative breast cancer microscopy images & Dice \\
    \bottomrule
  \end{tabular}
\end{table}

\clearpage

\begingroup
\bibliographystyle{plainnat}
\bibliography{mybibliography}

@article{wu2025unibiomed,
  title={A universal foundation model for grounded biomedical image interpretation},
  author={Wu, Linshan and Nie, Yuxiang and He, Sunan and Zhuang, Jiaxin and Luo, Luyang and Li, Tao and Xie, Zhuoyao and Chen, Dexuan and Zhao, Yinghua and Mahboobani, Neeraj and Vardhanabhuti, Varut and Chan, Ronald Cheong Kin and Peng, Yifan and Rajpurkar, Pranav and Chen, Hao},
  journal={Nature Communications},
  volume={17},
  number={1},
  pages={7173},
  year={2026},
  doi={10.1038/s41467-026-73986-1},
  note={\url{https://doi.org/10.1038/s41467-026-73986-1}}
}

@inproceedings{chen2025mimo,
  title={MIMO: A Medical Vision Language Model with Visual Referring Multimodal Input and Pixel Grounding Multimodal Output},
  author={Chen, Yanyuan and Xu, Dexuan and Huang, Yu and Zhan, Songkun and Wang, Hanpin and Chen, Dongxue and Wang, Xueping and Qiu, Meikang and Li, Hang},
  booktitle={Proceedings of the Computer Vision and Pattern Recognition Conference},
  pages={24732--24741},
  year={2025}
}

@inproceedings{lai2024lisa,
  title={Lisa: Reasoning segmentation via large language model},
  author={Lai, Xin and Tian, Zhuotao and Chen, Yukang and Li, Yanwei and Yuan, Yuhui and Liu, Shu and Jia, Jiaya},
  booktitle={Proceedings of the IEEE/CVF Conference on Computer Vision and Pattern Recognition},
  pages={9579--9589},
  year={2024}
}

@inproceedings{rasheed2024glamm,
  title={Glamm: Pixel grounding large multimodal model},
  author={Rasheed, Hanoona and Maaz, Muhammad and Shaji, Sahal and Shaker, Abdelrahman and Khan, Salman and Cholakkal, Hisham and others},
  booktitle={Proceedings of the IEEE Conference on Computer Vision and Pattern Recognition},
  pages={13009--13018},
  year={2024}
}

@inproceedings{hu2022lora,
  title={Lora: Low-rank adaptation of large language models.},
  author={Shen, Yelong and Wallis, Phillip and Allen-Zhu, Zeyuan and Li, Yuanzhi and Wang, Shean and others},
  booktitle={International Conference on Learning Representations},
  year={2024},
}

@inproceedings{kirillov2023segment,
  title={Segment anything},
  author={Kirillov, Alexander and Mintun, Eric and Ravi, Nikhila and Mao, Hanzi and Rolland, Chloe and Gustafson, Laura and Xiao, Tete and Whitehead, Spencer and Berg, Alexander C and Lo, Wan-Yen and others},
  booktitle={Proceedings of the IEEE/CVF international conference on computer vision},
  pages={4015--4026},
  year={2023}
}

@article{ravi2024sam,
  title={Sam 2: Segment anything in images and videos},
  author={Ravi, Nikhila and Gabeur, Valentin and Hu, Yuan-Ting and Hu, Ronghang and Ryali, Chaitanya and Ma, Tengyu and Khedr, Haitham and R{\"a}dle, Roman and Rolland, Chloe and Gustafson, Laura and others},
  journal={arXiv preprint arXiv:2408.00714},
  year={2024}
}

@article{huang2024towards,
  title={Towards a Multimodal Large Language Model with Pixel-Level Insight for Biomedicine},
  author={Huang, Xiaoshuang and Shen, Lingdong and Liu, Jia and Shang, Fangxin and Li, Hongxiang and Huang, Haifeng and Yang, Yehui},
  journal={arXiv preprint arXiv:2412.09278},
  year={2024}
}

@article{liu2023visual,
  title={Visual instruction tuning},
  author={Liu, Haotian and Li, Chunyuan and Wu, Qingyang and Lee, Yong Jae},
  journal={Advances in neural information processing systems},
  volume={36},
  pages={34892--34916},
  year={2023}
}

@article{yang2025qwen2.5,
  title={Qwen2.5-VL Technical Report},
  author={Shuai Bai and Keqin Chen and Xuejing Liu and Jialin Wang and Wenbin Ge and Sibo Song and others},
  journal={arXiv preprint arXiv:2502.13923},
  year={2025}
}

@inproceedings{rasley2020deepspeed,
  title={Deepspeed: System optimizations enable training deep learning models with over 100 billion parameters},
  author={Rasley, Jeff and Rajbhandari, Samyam and Ruwase, Olatunji and He, Yuxiong},
  booktitle={Proceedings of the 26th ACM SIGKDD international conference on knowledge discovery \& data mining},
  pages={3505--3506},
  year={2020}
}

@article{dao2023flashattention,
  title={Flashattention-2: Faster attention with better parallelism and work partitioning},
  author={Dao, Tri},
  journal={arXiv:2307.08691},
  year={2023}
}

@inproceedings{jin2019pubmedqa,
  title={Pubmedqa: A dataset for biomedical research question answering},
  author={Jin, Qiao and Dhingra, Bhuwan and Liu, Zhengping and Cohen, William and Lu, Xinghua},
  booktitle={Proceedings of the 2019 conference on empirical methods in natural language processing and the 9th international joint conference on natural language processing (EMNLP-IJCNLP)},
  pages={2567--2577},
  year={2019}
}

@inproceedings{pal2022medmcqa,
  title={Medmcqa: A large-scale multi-subject multi-choice dataset for medical domain question answering},
  author={Pal, Ankit and Umapathi, Logesh Kumar and Sankarasubbu, Malaikannan},
  booktitle={Conference on health, inference, and learning},
  pages={248--260},
  year={2022},
  organization={PMLR}
}

@article{jin2021disease,
  title={What disease does this patient have? a large-scale open domain question answering dataset from medical exams},
  author={Jin, Di and Pan, Eileen and Oufattole, Nassim and Weng, Wei-Hung and Fang, Hanyi and Szolovits, Peter},
  journal={Applied Sciences},
  volume={11},
  number={14},
  pages={6421},
  year={2021},
  publisher={MDPI}
}

@article{zuo2025medxpertqa,
  title={Medxpertqa: Benchmarking expert-level medical reasoning and understanding},
  author={Zuo, Yuxin and Qu, Shang and Li, Yifei and Chen, Zhangren and Zhu, Xuekai and Hua, Ermo and others},
  journal={arXiv:2501.18362},
  year={2025}
}

@inproceedings{li2024cmmlu,
  title={Cmmlu: Measuring massive multitask language understanding in chinese},
  author={Li, Haonan and Zhang, Yixuan and Koto, Fajri and Yang, Yifei and Zhao, Hai and Gong, Yeyun and others},
  booktitle={Findings of the Association for Computational Linguistics: ACL 2024},
  pages={11260--11285},
  year={2024}
}

@article{lau2018dataset,
  title={A dataset of clinically generated visual questions and answers about radiology images},
  author={Lau, Jason J and Gayen, Soumya and Ben Abacha, Asma and Demner-Fushman, Dina},
  journal={Scientific data},
  volume={5},
  number={1},
  pages={180251},
  year={2018},
  publisher={Nature Publishing Group}
}

@inproceedings{liu2021slake,
  title={Slake: A semantically-labeled knowledge-enhanced dataset for medical visual question answering},
  author={Liu, Bo and Zhan, Li-Ming and Xu, Li and Ma, Lin and Yang, Yan and Wu, Xiao-Ming},
  booktitle={2021 18th international symposium on biomedical imaging},
  pages={1650--1654},
  year={2021},
  organization={IEEE}
}

@article{naseem2022vision,
  title={Vision-language transformer for interpretable pathology visual question answering},
  author={Naseem, Usman and Khushi, Matloob and Kim, Jinman},
  journal={IEEE journal of biomedical and health informatics},
  volume={27},
  number={4},
  pages={1681--1690},
  year={2022},
  publisher={IEEE}
}

@article{zhang2023pmc,
  title={Pmcvqa: Visual instruction tuning for medical visual question answering},
  author={Zhang, Xiaoman and Wu, Chaoyi and Zhao, Ziheng and Lin, Weixiong and Zhang, Ya and Wang, Yanfeng and others},
  journal={arXiv:2305.10415},
  year={2023}
}

@article{wang2025citrus,
  title={Citrus-V: Advancing Medical Foundation Models with Unified Medical Image Grounding for Clinical Reasoning},
  author={Wang, Guoxin and Zhao, Jun and Liu, Xinyi and Liu, Yanbo and Cao, Xuyang and Li, Chao and others},
  journal={arXiv:2509.19090},
  year={2025}
}

@article{kucs2024medsegbench,
  title={MedSegBench: A comprehensive benchmark for medical image segmentation in diverse data modalities},
  author={Ku{\c{s}}, Zeki and Aydin, Musa},
  journal={Scientific Data},
  volume={11},
  number={1},
  pages={1283},
  year={2024},
  publisher={Nature Publishing Group UK London}
}

@article{xu2025lingshu,
  title={Lingshu: A generalist foundation model for unified multimodal medical understanding and reasoning},
  author={Xu, Weiwen and Chan, Hou Pong and Li, Long and Aljunied, Mahani and Yuan, Ruifeng and Wang, Jianyu and others},
  journal={arXiv:2506.07044},
  year={2025}
}

@inproceedings{lin2025healthgpt,
  title={HealthGPT: A Medical Large Vision-Language Model for Unifying Comprehension and Generation via Heterogeneous Knowledge Adaptation},
  author={Lin, Tianwei and Zhang, Wenqiao and Li, Sijing and Yuan, Yuqian and Yu, Binhe and Li, Haoyuan and others},
  booktitle={International Conference on Machine Learning},
  pages={37975--37995},
  year={2025},
  organization={PMLR}
}

@inproceedings{chen2024towards,
  title={Towards injecting medical visual knowledge into multimodal llms at scale},
  author={Chen, Junying and Gui, Chi and Ouyang, Ruyi and Gao, Anningzhe and Chen, Shunian and Chen, Guiming Hardy and others},
  booktitle={Proceedings of the 2024 conference on empirical methods in natural language processing},
  pages={7346--7370},
  year={2024}
}

@article{sellergren2025medgemma,
  title={Medgemma technical report},
  author={Sellergren, Andrew and Kazemzadeh, Sahar and Jaroensri, Tiam and Kiraly, Atilla and Traverse, Madeleine and Kohlberger, Timo and others},
  journal={arXiv:2507.05201},
  year={2025}
}

@inproceedings{ronneberger2015u,
  title={U-net: Convolutional networks for biomedical image segmentation},
  author={Ronneberger, Olaf and Fischer, Philipp and Brox, Thomas},
  booktitle={International Conference on Medical image computing and computer-assisted intervention},
  pages={234--241},
  year={2015},
  organization={Springer}
}

@inproceedings{tan2019efficientnet,
  title={Efficientnet: Rethinking model scaling for convolutional neural networks},
  author={Tan, Mingxing and Le, Quoc},
  booktitle={ICML},
  pages={6105--6114},
  year={2019},
  organization={PMLR}
}

@article{iandola2014densenet,
  title={Densenet: Implementing efficient convnet descriptor pyramids},
  author={Iandola, Forrest and Moskewicz, Matt and Karayev, Sergey and Girshick, Ross and Darrell, Trevor and Keutzer, Kurt},
  journal={arXiv:1404.1869},
  year={2014}
}

@inproceedings{he2016deep,
  title={Deep residual learning for image recognition},
  author={He, Kaiming and Zhang, Xiangyu and Ren, Shaoqing and Sun, Jian},
  booktitle={Proceedings of the IEEE conference on computer vision and pattern recognition},
  pages={770--778},
  year={2016}
}

@inproceedings{sandler2018mobilenetv2,
  title={Mobilenetv2: Inverted residuals and linear bottlenecks},
  author={Sandler, Mark and Howard, Andrew and Zhu, Menglong and Zhmoginov, Andrey and Chen, Liang-Chieh},
  booktitle={Proceedings of the IEEE conference on computer vision and pattern recognition},
  pages={4510--4520},
  year={2018}
}

@article{van2008visualizing,
  title={Visualizing data using t-SNE.},
  author={Van der Maaten, Laurens and Hinton, Geoffrey},
  journal={Journal of machine learning research},
  volume={9},
  number={11},
  year={2008}
}

@article{davies2009cluster,
  title={A cluster separation measure},
  author={Davies, David L and Bouldin, Donald W},
  journal={IEEE transactions on pattern analysis and machine intelligence},
  number={2},
  pages={224--227},
  year={2009},
  publisher={Ieee}
}

@inproceedings{chen2018gradnorm,
  title={GradNorm: Gradient normalization for adaptive loss balancing in deep multitask networks},
  author={Chen, Zhao and Badrinarayanan, Vijay and Lee, Chen-Yu and Rabinovich, Andrew},
  booktitle={International conference on machine learning},
  year={2018}
}

@inproceedings{yu2020gradient,
  title={Gradient surgery for multi-task learning},
  author={Yu, Tianhe and Kumar, Saurabh and Gupta, Abhishek and Levine, Sergey and Hausman, Karol and Finn, Chelsea},
  booktitle={Advances in Neural Information Processing Systems},
  year={2020}
}

@article{ma2024segment,
  title={Segment anything in medical images},
  author={Ma, Jun and He, Yuting and Li, Feifei and Han, Lin and You, Chenyu and Wang, Bo},
  journal={Nature Communications},
  year={2024}
}

@inproceedings{li2023llavamed,
  title={LLaVA-Med: Training a Large Language-and-Vision Assistant for Biomedicine in One Day},
  author={Li, Chunyuan and Wong, Cliff and Zhang, Sheng and Usuyama, Naoto and Liu, Haotian and Yang, Jianwei and Naumann, Tristan and Poon, Hoifung and Gao, Jianfeng},
  booktitle={Advances in Neural Information Processing Systems},
  volume={36},
  pages={28541--28564},
  year={2023}
}

@article{wu2023radfm,
  title={Towards Generalist Foundation Model for Radiology by Leveraging Web-scale 2D and 3D Medical Data},
  author={Wu, Chaoyi and Zhang, Xiaoman and Zhang, Ya and Wang, Yanfeng and Xie, Weidi},
  journal={Nature Communications},
  volume={16},
  year={2025},
  doi={10.1038/s41467-025-62385-7}
}

@article{wu2024pmcllama,
  title={PMC-LLaMA: toward building open-source language models for medicine},
  author={Wu, Chaoyi and Lin, Weixiong and Zhang, Xiaoman and Zhang, Ya and Xie, Weidi and Wang, Yanfeng},
  journal={Journal of the American Medical Informatics Association},
  volume={31},
  number={9},
  pages={1833--1843},
  year={2024},
  doi={10.1093/jamia/ocae045}
}

@inproceedings{ren2024pixellm,
  title={PixelLM: Pixel Reasoning with Large Multimodal Model},
  author={Ren, Zhongwei and Huang, Zhicheng and Wei, Yunchao and Zhao, Yao and Fu, Dongmei and Feng, Jiashi and Jin, Xiaojie},
  booktitle={Proceedings of the IEEE/CVF Conference on Computer Vision and Pattern Recognition},
  pages={26364--26373},
  year={2024}
}

@inproceedings{yuan2024osprey,
  title={Osprey: Pixel Understanding with Visual Instruction Tuning},
  author={Yuan, Yuqian and Li, Wentong and Liu, Jian and Tang, Dongqi and Luo, Xinjie and Qin, Chi and Zhang, Lei and Zhu, Jianke},
  booktitle={Proceedings of the IEEE/CVF Conference on Computer Vision and Pattern Recognition},
  pages={28202--28211},
  year={2024}
}

@inproceedings{zou2023seem,
  title={Segment Everything Everywhere All at Once},
  author={Zou, Xueyan and Yang, Jianwei and Zhang, Hao and Li, Feng and Li, Linjie and Wang, Jianfeng and Wang, Lijuan and Gao, Jianfeng and Lee, Yong Jae},
  booktitle={Advances in Neural Information Processing Systems},
  volume={36},
  year={2023}
}

@inproceedings{liu2024groundingdino,
  title={Grounding DINO: Marrying DINO with Grounded Pre-training for Open-Set Object Detection},
  author={Liu, Shilong and Zeng, Zhaoyang and Ren, Tianhe and Li, Feng and Zhang, Hao and Yang, Jie and Li, Chunyuan and Yang, Jianwei and Su, Hang and Zhu, Jun and Zhang, Lei},
  booktitle={European Conference on Computer Vision},
  series={Lecture Notes in Computer Science},
  pages={38--55},
  year={2024},
  doi={10.1007/978-3-031-72970-6_3}
}

@inproceedings{navon2022nashmtl,
  title={Multi-Task Learning as a Bargaining Game},
  author={Navon, Aviv and Shamsian, Aviv and Achituve, Idan and Maron, Haggai and Kawaguchi, Kenji and Chechik, Gal and Fetaya, Ethan},
  booktitle={Proceedings of the 39th International Conference on Machine Learning},
  series={Proceedings of Machine Learning Research},
  volume={162},
  year={2022},
  organization={PMLR}
}

@inproceedings{sener2018mgda,
  title={Multi-Task Learning as Multi-Objective Optimization},
  author={Sener, Ozan and Koltun, Vladlen},
  booktitle={Advances in Neural Information Processing Systems},
  volume={31},
  year={2018}
}

@inproceedings{kendall2018multitask,
  title={Multi-task learning using uncertainty to weigh losses for scene geometry and semantics},
  author={Kendall, Alex and Gal, Yarin and Cipolla, Roberto},
  booktitle={Proceedings of the IEEE Conference on Computer Vision and Pattern Recognition},
  pages={7482--7491},
  year={2018}
}

@inproceedings{radford2021clip,
  title={Learning Transferable Visual Models From Natural Language Supervision},
  author={Radford, Alec and Kim, Jong Wook and Hallacy, Chris and Ramesh, Aditya and Goh, Gabriel and Agarwal, Sandhini and Sastry, Girish and Askell, Amanda and Mishkin, Pamela and Clark, Jack and others},
  booktitle={Proceedings of the 38th International Conference on Machine Learning},
  series={Proceedings of Machine Learning Research},
  volume={139},
  year={2021},
  organization={PMLR}
}

@inproceedings{pan2025dusss,
  title={{DuSSS}: Dual Semantic Similarity-Supervised Vision-Language Model for Semi-Supervised Medical Image Segmentation},
  author={Pan, Qingtao and Qiao, Wenhao and Lou, Jingjiao and Ji, Bing and Li, Shuo},
  booktitle={Proceedings of the AAAI Conference on Artificial Intelligence},
  volume={39},
  pages={6299--6307},
  year={2025},
  doi={10.1609/aaai.v39i6.32674}
}

@inproceedings{polamreddy2026cgreasoner,
  author={Polamreddy, Lakshmikar Reddy and Ma, Ming},
  title={{CG-Reasoner}: Centroid-Guided Positional Reasoning Segmentation for Medical Imaging with a Robust Visual-Text Consistency Metric},
  booktitle={Proceedings of the IEEE/CVF Conference on Computer Vision and Pattern Recognition},
  pages={1472--1481},
  year={2026}
}

@inproceedings{zhang2026medkco,
  author={Zhang, Chenran and Wu, Ruiqi and Zhou, Tao and Zhou, Yi},
  title={{MedKCO}: Medical Vision-Language Pretraining via Knowledge-Driven Cognitive Orchestration},
  booktitle={Proceedings of the IEEE/CVF Conference on Computer Vision and Pattern Recognition},
  pages={35260--35269},
  year={2026}
}

@inproceedings{nejatimanzari2026sparse,
  author={Nejatimanzari, Omid and Asgariandehkordi, Hojat and Koleilat, Taha and Xiao, Yiming and Rivaz, Hassan},
  title={Sparse Spectral {LoRA}: Routed Experts for Medical {VLMs}},
  booktitle={Proceedings of the IEEE/CVF Conference on Computer Vision and Pattern Recognition},
  pages={35351--35362},
  year={2026}
}

@article{johnson2019mimiccxr,
  title={{MIMIC-CXR}, a de-identified publicly available database of chest radiographs with free-text reports},
  author={Johnson, Alistair E. W. and Pollard, Tom J. and Berkowitz, Seth J. and Greenbaum, Nathaniel R. and Lungren, Matthew P. and Deng, Chih-ying and Mark, Roger G. and Horng, Steven},
  journal={Scientific Data},
  volume={6},
  number={1},
  pages={317},
  year={2019}
}

@article{chen2023huatuogptii,
  title={{HuatuoGPT-II}, One-stage Training for Medical Adaption of {LLMs}},
  author={Chen, Junying and Wang, Xiaohan and Gao, Hongbo and Li, Feng and Jiang, Xiangwan and Tang, Huaizhi and Ding, Benyou and Zhou, Jing and Zhang, Yuchen and Qi, Zhaoxiang and others},
  journal={arXiv preprint arXiv:2311.09774},
  year={2023}
}

@article{chen2024allava,
  title={{ALLaVA}: Harnessing {GPT4V}-Synthesized Data for a Lite Vision-Language Model},
  author={Chen, Guo and Shen, Liao and Shao, Rui and Deng, Xiang and Nie, Liqiang},
  journal={arXiv preprint arXiv:2402.11684},
  year={2024}
}

@misc{lin2023pmcclipcontrastivelanguageimagepretraining,
      title={PMC-CLIP: Contrastive Language-Image Pre-training using Biomedical Documents}, 
      author={Weixiong Lin and Ziheng Zhao and Xiaoman Zhang and Chaoyi Wu and Ya Zhang and Yanfeng Wang and Weidi Xie},
      year={2023},
      eprint={2303.07240},
      archivePrefix={arXiv},
      primaryClass={cs.CV},
      url={https://arxiv.org/abs/2303.07240}, 
}

@InProceedings{ROCO,
author="Pelka, Obioma
and Koitka, Sven
and R{\"u}ckert, Johannes
and Nensa, Felix
and Friedrich, Christoph M.",
editor="Stoyanov, Danail
and Taylor, Zeike
and Balocco, Simone
and Sznitman, Raphael
and Martel, Anne
and Maier-Hein, Lena
and Duong, Luc
and Zahnd, Guillaume
and Demirci, Stefanie
and Albarqouni, Shadi
and Lee, Su-Lin
and Moriconi, Stefano
and Cheplygina, Veronika
and Mateus, Diana
and Trucco, Emanuele
and Granger, Eric
and Jannin, Pierre",
title="Radiology Objects in COntext (ROCO): A Multimodal Image Dataset",
booktitle="Intravascular Imaging and Computer Assisted Stenting and Large-Scale Annotation of Biomedical Data and Expert Label Synthesis",
year="2018",
publisher="Springer International Publishing",
address="Cham",
pages="180--189",
isbn="978-3-030-01364-6"
}

@article{li2023llava,
  title={Llava-med: Training a large language-and-vision assistant for biomedicine in one day},
  author={Li, Chunyuan and Wong, Cliff and Zhang, Sheng and Usuyama, Naoto and Liu, Haotian and Yang, Jianwei and Naumann, Tristan and Poon, Hoifung and Gao, Jianfeng},
  journal={Advances in Neural Information Processing Systems},
  volume={36},
  pages={28541--28564},
  year={2023}
}

@article{chambon2024chexpertplusaugmentinglarge,
      title={CheXpert Plus: Augmenting a Large Chest X-ray Dataset with Text Radiology Reports, Patient Demographics and Additional Image Formats}, 
      author={Pierre Chambon and Jean-Benoit Delbrouck and Thomas Sounack and Shih-Cheng Huang and Zhihong Chen and Maya Varma and Steven QH Truong and Chu The Chuong and Curtis P. Langlotz},
      year={2024},
      volume={abs/2405.19538},
      journal={ArXiv},
      url={https://arxiv.org/abs/2405.19538}, 
}

@article{ROCOv2,
	author = {R{\"u}ckert, Johannes and Bloch, Louise and Br{\"u}ngel, Raphael and Idrissi-Yaghir, Ahmad and Sch{\"a}fer, Henning and Schmidt, Cynthia S. and Koitka, Sven and Pelka, Obioma and Abacha, Asma Ben and G. Seco de Herrera, Alba and M{\"u}ller, Henning and Horn, Peter A. and Nensa, Felix and Friedrich, Christoph M.},
	date = {2024/06/26},
	doi = {10.1038/s41597-024-03496-6},
	id = {R{\"u}ckert2024},
	isbn = {2052-4463},
	journal = {Scientific Data},
	number = {1},
	pages = {688},
	title = {ROCOv2: Radiology Objects in COntext Version 2, an Updated Multimodal Image Dataset},
	url = {https://doi.org/10.1038/s41597-024-03496-6},
	volume = {11},
	year = {2024}}

@article{chen2024huatuogptvision,
      title={HuatuoGPT-Vision, Towards Injecting Medical Visual Knowledge into Multimodal LLMs at Scale}, 
      author={Junying Chen and Chi Gui and Ruyi Ouyang and Anningzhe Gao and Shunian Chen and Guiming Hardy Chen and Xidong Wang and Ruifei Zhang and Zhenyang Cai and Ke Ji and Guangjun Yu and Xiang Wan and Benyou Wang},
      year={2024},
      volume={abs/2406.19280},
      journal={ArXiv},
      url={https://arxiv.org/abs/2406.19280}, 
}

@article{demner2016preparing-iu-xray,
  title={Preparing a collection of radiology examinations for distribution and retrieval},
  author={Demner-Fushman, Dina and Kohli, Marc D and Rosenman, Marc B and Shooshan, Sonya E and Rodriguez, Laritza and Antani, Sameer and Thoma, George R and McDonald, Clement J},
  journal={Journal of the American Medical Informatics Association},
  volume={23},
  number={2},
  pages={304--310},
  year={2016},
  publisher={Oxford University Press},
  url={https://doi.org/10.1093/jamia/ocv080}
}

@article{bae2024ehrxqa,
  title={EHRXQA: A multi-modal question answering dataset for electronic health records with chest x-ray images},
  author={Bae, Seongsu and Kyung, Daeun and Ryu, Jaehee and Cho, Eunbyeol and Lee, Gyubok and Kweon, Sunjun and Oh, Jungwoo and Ji, Lei and Chang, Eric and Kim, Tackeun and others},
  journal={Advances in Neural Information Processing Systems},
  volume={36},
  year={2024}
}

@Inproceedings{ImageCLEFVQA-Med2019,
  author = {Asma {Ben Abacha} and Sadid A. Hasan and Vivek V. Datla and Joey Liu and Dina Demner-Fushman and Henning M\"uller},
  title = {VQA-Med: Overview of the Medical Visual Question Answering Task at ImageCLEF 2019},
  url = {https://ceur-ws.org/Vol-2380/paper\_272.pdf},
  booktitle = {Working Notes of {CLEF} 2019},
  series = {{CEUR} Workshop Proceedings},
  volume = {2380},
  year = {2019},
  publisher    = {CEUR-WS.org}, 
  month = {September 9-12},
  address = {Lugano, Switzerland}
}

@misc{wang2025citrusleveragingexpertcognitive,
      title={Citrus: Leveraging Expert Cognitive Pathways in a Medical Language Model for Advanced Medical Decision Support}, 
      author={Guoxin Wang and Minyu Gao and Shuai Yang and Ya Zhang and Lizhi He and Liang Huang and Hanlin Xiao and Yexuan Zhang and Wanyue Li and Lu Chen and Jintao Fei and Xin Li},
      year={2025},
      eprint={2502.18274},
      archivePrefix={arXiv},
      primaryClass={cs.AI},
      url={https://arxiv.org/abs/2502.18274}, 
}

@misc{li2023chatdoctormedicalchatmodel,
      title={ChatDoctor: A Medical Chat Model Fine-Tuned on a Large Language Model Meta-AI (LLaMA) Using Medical Domain Knowledge}, 
      author={Yunxiang Li and Zihan Li and Kai Zhang and Ruilong Dan and Steve Jiang and You Zhang},
      year={2023},
      eprint={2303.14070},
      archivePrefix={arXiv},
      primaryClass={cs.CL},
      url={https://arxiv.org/abs/2303.14070}, 
}

@misc{zhang2025alpacareinstructiontunedlargelanguagemodels,
      title={AlpaCare:Instruction-tuned Large Language Models for Medical Application}, 
      author={Xinlu Zhang and Chenxin Tian and Xianjun Yang and Lichang Chen and Zekun Li and Linda Ruth Petzold},
      year={2025},
      eprint={2310.14558},
      archivePrefix={arXiv},
      primaryClass={cs.CL},
      url={https://arxiv.org/abs/2310.14558}, 
}

@misc{wang2024apollo,
   title={Apollo: Lightweight Multilingual Medical LLMs towards Democratizing Medical AI to 6B People},
   author={Xidong Wang and Nuo Chen and Junyin Chen and Yan Hu and Yidong Wang and Xiangbo Wu and Anningzhe Gao and Xiang Wan and Haizhou Li and Benyou Wang},
   year={2024},
   eprint={2403.03640},
   archivePrefix={arXiv},
   primaryClass={cs.CL}
}

@ARTICLE{BenAbacha-BMC-2019,    
    author = {Asma {Ben Abacha} and Dina Demner{-}Fushman},
    title = {A Question-Entailment Approach to Question Answering},
    journal = {{BMC} Bioinform.}, 
    volume = {20},
      number = {1},
        pages = {511:1--511:23},
      year = {2019},
  url = {https://bmcbioinformatics.biomedcentral.com/articles/10.1186/s12859-019-3119-4}
}

@misc{wu2023pmcllamabuildingopensourcelanguage,
      title={PMC-LLaMA: Towards Building Open-source Language Models for Medicine}, 
      author={Chaoyi Wu and Weixiong Lin and Xiaoman Zhang and Ya Zhang and Yanfeng Wang and Weidi Xie},
      year={2023},
      eprint={2304.14454},
      archivePrefix={arXiv},
      primaryClass={cs.CL},
      url={https://arxiv.org/abs/2304.14454}, 
}

@misc{chen2024huatuogpto1medicalcomplexreasoning,
      title={HuatuoGPT-o1, Towards Medical Complex Reasoning with LLMs}, 
      author={Junying Chen and Zhenyang Cai and Ke Ji and Xidong Wang and Wanlong Liu and Rongsheng Wang and Jianye Hou and Benyou Wang},
      year={2024},
      eprint={2412.18925},
      archivePrefix={arXiv},
      primaryClass={cs.CL},
      url={https://arxiv.org/abs/2412.18925}, 
}

@misc{wu2025medreasonelicitingfactualmedical,
      title={MedReason: Eliciting Factual Medical Reasoning Steps in LLMs via Knowledge Graphs}, 
      author={Juncheng Wu and Wenlong Deng and Xingxuan Li and Sheng Liu and Taomian Mi and Yifan Peng and Ziyang Xu and Yi Liu and Hyunjin Cho and Chang-In Choi and Yihan Cao and Hui Ren and Xiang Li and Xiaoxiao Li and Yuyin Zhou},
      year={2025},
      eprint={2504.00993},
      archivePrefix={arXiv},
      primaryClass={cs.CL},
      url={https://arxiv.org/abs/2504.00993}, 
}

@misc{medthoughts8k,
  title = {MedThoughts-8K Dataset},
  author = {hw-hwei},
  year = {2025},
  publisher = {HuggingFace},
  url = {https://huggingface.co/datasets/hw-hwei/MedThoughts-8K}
}

@misc{liu2024improvedbaselinesvisualinstruction,
      title={Improved Baselines with Visual Instruction Tuning}, 
      author={Haotian Liu and Chunyuan Li and Yuheng Li and Yong Jae Lee},
      year={2024},
      eprint={2310.03744},
      archivePrefix={arXiv},
      primaryClass={cs.CV},
      url={https://arxiv.org/abs/2310.03744}, 
}

@misc{deitke2024molmopixmoopenweights,
      title={Molmo and PixMo: Open Weights and Open Data for State-of-the-Art Vision-Language Models}, 
      author={Matt Deitke and Christopher Clark and Sangho Lee and others},
      year={2024},
      eprint={2409.17146},
      archivePrefix={arXiv},
      primaryClass={cs.CV},
      url={https://arxiv.org/abs/2409.17146}, 
}

@misc{OpenHermes2-5,
  title = {OpenHermes 2.5: An Open Dataset of Synthetic Data for Generalist LLM Assistants},
  author = {Teknium},
  year = {2023},
  publisher = {HuggingFace},
  url = {https://huggingface.co/datasets/teknium/OpenHermes-2.5}
}

@misc{marino2019okvqavisualquestionanswering,
      title={OK-VQA: A Visual Question Answering Benchmark Requiring External Knowledge}, 
      author={Kenneth Marino and Mohammad Rastegari and Ali Farhadi and Roozbeh Mottaghi},
      year={2019},
      eprint={1906.00067},
      archivePrefix={arXiv},
      primaryClass={cs.CV},
      url={https://arxiv.org/abs/1906.00067}, 
}

@misc{schwenk2022aokvqabenchmarkvisualquestion,
      title={A-OKVQA: A Benchmark for Visual Question Answering using World Knowledge}, 
      author={Dustin Schwenk and Apoorv Khandelwal and Christopher Clark and Kenneth Marino and Roozbeh Mottaghi},
      year={2022},
      eprint={2206.01718},
      archivePrefix={arXiv},
      primaryClass={cs.CV},
      url={https://arxiv.org/abs/2206.01718}, 
}

@InProceedings{mishraICDAR19,
  author    = "Anand Mishra and Shashank Shekhar and Ajeet Kumar Singh and Anirban Chakraborty",
  title     = "OCR-VQA: Visual Question Answering by Reading Text in Images",
  booktitle = "ICDAR",
  year      = "2019",
}

@InProceedings{sidorov2019textcaps,
    title={TextCaps: a Dataset for Image Captioningwith Reading Comprehension},
    author={Sidorov, Oleksii and Hu, Ronghang and Rohrbach, Marcus and Singh, Amanpreet},
    booktitle={European Conference on Computer Vision},
    year={2020}
}

@article{siragusa2025medpix20,
  title={MedPix 2.0: A Comprehensive Multimodal Biomedical Dataset for Advanced AI Applications},
  author={Siragusa, Enrico and Gessi, Giovanni and Schettini, Francesco and Renda, Maria Elena and Gambella, Alessandro and Gentili, Claudio and Morelli, Laura and Zuppardo, Luca and Curti, Nicolas and Coppola, Francesca and others},
  journal={arXiv preprint arXiv:2507.02994},
  year={2025}
}

@article{seyfioglu2024quiltllava,
  title={Quilt-LLaVA: Visual Instruction Tuning by Extracting Localized Narratives from Open-Source Histopathology Videos},
  author={Seyfioglu, Mehmet Saygin and Ikezogwo, Wisdom and Ghezloo, Farhad and Krishna, Ranjay and Shapiro, Linda},
  journal={Proceedings of the IEEE/CVF Conference on Computer Vision and Pattern Recognition Workshops},
  year={2024}
}

@inproceedings{Isic2016MSBench,
  title = {Skin Lesion Analysis Toward Melanoma Detection: A Challenge at the 2017 International Symposium on Biomedical Imaging (ISBI), Hosted by the International Skin Imaging Collaboration (ISIC)},
  author = {Codella, Noel C. F. and Gutman, David and Celebi, M. Emre and Helba, Brian and Marchetti, Michael A. and Dusza, Stephen W. and Kalloo, Aadi and Liopyris, Konstantinos and Mishra, Nabin and Kittler, Harald and Halpern, Allan},
  booktitle = {2018 IEEE 15th International Symposium on Biomedical Imaging (ISBI 2018)},
  pages = {168--172},
  year = {2018},
  organization = {IEEE},
  doi = {10.1109/ISBI.2018.8363547},
  url = {https://doi.org/10.1109/ISBI.2018.8363547}
}

@incollection{KvasirMSBench,
  title = {Kvasir-SEG: A Segmented Polyp Dataset},
  author = {Jha, Debesh and Smedsrud, Pia H. and Riegler, Michael A. and Halvorsen, P{\aa}l and de Lange, Thomas and Johansen, Dag and Johansen, H{\aa}vard D.},
  booktitle = {MultiMedia Modeling},
  pages = {451--462},
  year = {2020},
  publisher = {Springer International Publishing},
  doi = {10.1007/978-3-030-37734-2_37},
  url = {https://doi.org/10.1007/978-3-030-37734-2_37}
}

@article{IdribMSBench,
  title = {Indian Diabetic Retinopathy Image Dataset (IDRiD): A Database for Diabetic Retinopathy Screening Research},
  author = {Porwal, Prasanna and Pachade, Samiksha and Kamble, Ravi and Kokare, Manesh and Deshmukh, Girish and Sahasrabuddhe, Vivek and Meriaudeau, Fabrice},
  journal = {Data},
  volume = {3},
  number = {3},
  pages = {25},
  year = {2018},
  publisher = {MDPI},
  doi = {10.3390/data3030025},
  url = {https://doi.org/10.3390/data3030025}
}

@article{CovidQUExMSBench,
  title = {COVID-19 Infection Localization and Severity Grading from Chest X-Ray Images},
  author = {Tahir, Anas M. and Chowdhury, Muhammad E. H. and Khandakar, Amith and Rahman, Tawsifur and Qiblawey, Yazan and Khurshid, Uzair and Kiranyaz, Serkan and Ibtehaz, Nabil and Rahman, M. Sohel and Al-Maadeed, Somaya and Mahmud, Sakib and Ezeddin, Maymouna and Hameed, Khaled and Hamid, Tahir},
  journal = {Computers in Biology and Medicine},
  volume = {139},
  pages = {105002},
  year = {2021},
  publisher = {Elsevier},
  doi = {10.1016/j.compbiomed.2021.105002},
  url = {https://doi.org/10.1016/j.compbiomed.2021.105002}
}

@article{Promise12MSBench,
  title = {Evaluation of Prostate Segmentation Algorithms for MRI: The PROMISE12 Challenge},
  author = {Litjens, Geert and Toth, Robert and van de Ven, Wendy and Hoeks, Caroline and Kerkstra, Sjoerd and van Ginneken, Bram and Vincent, Graham and Guillard, Gwenael and Birbeck, Neil and Zhang, Jindang and Strand, Robin and Malmberg, Filip and Ou, Yangming and Davatzikos, Christos and Kirschner, Matthias and Jung, Florian and Yuan, Jing and Qiu, Wu and Gao, Qinquan and Edwards, Philip and Maan, Bianca and van der Heijden, Ferdinand and Ghose, Soumya and Mitra, Jhimli and Dowling, Jason and Barratt, Dean and Huisman, Henkjan and Madabhushi, Anant},
  journal = {Medical Image Analysis},
  volume = {18},
  number = {2},
  pages = {359--373},
  year = {2014},
  publisher = {Elsevier},
  doi = {10.1016/j.media.2013.12.002},
  url = {https://doi.org/10.1016/j.media.2013.12.002}
}

@misc{MosMedPlusMSBench,
  title = {MosMedData: Chest CT Scans With COVID-19 Related Findings Dataset},
  author = {Morozov, S. P. and Andreychenko, A. E. and Pavlov, N. A. and Vladzymyrskyy, A. V. and Ledikhova, N. V. and Gombolevskiy, V. A. and Blokhin, I. A. and Gelezhe, P. B. and Gonchar, A. V. and Chernina, V. Yu.},
  year = {2020},
  eprint = {2005.06465},
  archivePrefix = {arXiv},
  url = {https://arxiv.org/abs/2005.06465}
}

@misc{UltrasoundNerveMSBench,
  title = {Ultrasound Nerve Segmentation},
  author = {{Kaggle}},
  year = {n.d.},
  howpublished = {\url{https://www.kaggle.com/competitions/ultrasound-nerve-segmentation}},
  note = {Accessed: 2026-07-30}
}

@misc{TnbcnucleiMSBench,
  title = {Segmentation of Nuclei in Histopathology Images by Deep Regression of the Distance Map},
  author = {Naylor, Peter Jack and Walter, Thomas and La{\'e}, Marick and Reyal, Fabien},
  year = {2018},
  publisher = {Zenodo},
  doi = {10.5281/zenodo.1175282},
  url = {https://doi.org/10.5281/zenodo.1175282}
}

@misc{ye2023samed2d20m,
  title = {SA-Med2D-20M Dataset: Segment Anything in 2D Medical Imaging with 20 Million Masks},
  author = {Ye, Jin and others},
  year = {2023},
  eprint = {2311.11969},
  archivePrefix = {arXiv},
  primaryClass = {cs.CV},
  url = {https://arxiv.org/abs/2311.11969}
}
\endgroup

\end{document}